\pdfoutput=1

\documentclass[10pt,twocolumn,letterpaper]{article}
\usepackage{cvpr}

\usepackage[accsupp]{axessibility} 
\usepackage{times}
\usepackage{epsfig}
\usepackage{graphicx}
\usepackage{amsmath}
\usepackage{amssymb}
\usepackage{dsfont}
\usepackage{pifont}% http://ctan.org/pkg/pifont
\usepackage{enumitem}
\usepackage{xcolor}
\usepackage{amsfonts}
\usepackage{dsfont}

\usepackage{soul}
\usepackage{subfigure}
\usepackage{amsfonts}
\usepackage{amsthm}
\usepackage{booktabs}
\usepackage{arydshln}

\usepackage[subtle]{savetrees}

\usepackage{xurl}
\usepackage{color, colortbl}
\definecolor{LightCyan}{rgb}{0.88,1,1}
\newcommand{\comment}[1]{}

\cvprfinalcopy % *** Uncomment this line for the final submission

\def\cvprPaperID{4857}

\ifcvprfinal\fi
\title{Learning Spatial-Temporal Graphs for Active Speaker Detection}
\begin{document}

\author{Sourya Roy$^{1\dagger}$\thanks{Work partially done during an internship at Intel Labs.} \quad\quad Kyle Min$^2$\thanks{Authors contributed equally.} \quad\quad Subarna Tripathi$^2$ \quad\quad Tanaya Guha$^3$ \quad\quad Somdeb Majumdar$^2$\\
$^1$UC Riverside \quad\quad $^2$Intel Labs \quad\quad $^3$University of Glasgow\\
{\tt\small sroy004@ucr.edu, \{kyle.min, subarna.tripathi, somdeb.majumdar\}@intel.com, tanaya.guha@glasgow.ac.uk}
}
\maketitle

\label{abs}
\begin{abstract}
We address the problem of active speaker detection through a new framework, called SPELL, that learns long-range multimodal graphs to encode the inter-modal relationship between audio and visual data. We cast active speaker detection as a node classification task that is aware of longer-term dependencies. We first construct a graph from a video so that each node corresponds to one person. 
Nodes representing the same identity share edges between them within a defined temporal window. Nodes within the same video frame are also connected to encode inter-person interactions. Through extensive experiments on the  Ava-ActiveSpeaker dataset, we demonstrate that learning graph-based representation, owing to its explicit spatial and temporal structure, significantly improves the overall performance. SPELL outperforms several relevant baselines and performs at par with state of the art models while requiring an order of magnitude lower computation cost. %The nodes in our graph has visual and audio features, and we pose the active speaker detection as a node classification problem considering longer-term dependencies. We first embed a graph on video in a canonical way so that nodes with similar identity over time share edges between them for a defined temporal window. Additionally nodes having same time-stamp are also connected to each other. 
% Next, we propose how to train our model effectively 
% in a self-supervised way utilizing 
% a pretext task of audio-video sync prediction and a
% task-specific data augmentation. 
%We demonstrate that learning graph-based representation, owing to its explicit spatial and temporal structure, significantly improves the representation of node features over a baseline of pretrained dual stream network, with negligible parameters overhead. We show the efficacy of our method on Ava-ActiveSpeaker dataset. 
%SPELL performs comparably with SOTA models, yet requiring one order of magnitude lower computation cost.

% 

\end{abstract}

\section{Introduction}
\label{sec:intro}
%Computer vision has made tremendous progress over last two decades across a large array of problems. In many traditional tasks such as objection detection, face recognition etc. near human level performance have been achieved and a large part of this success can be attributed towards deep learning based models. %However, 
%\tanaya{The sentences above are too generic. The intro is a bit verbose, and has paragraphs too long which is not good for readability. Sorry, I am editing.}
Holistic scene understanding in its full generality is still a challenge in computer vision despite breakthroughs in several other areas. A \emph{scene} represents real-life events spanning complex visual and auditory information, often intertwined.
%still seems like riddled with challenges. One key aspect of scene understanding that has not been taken into consideration until very recently is its multi-modal nature. %images, sounds, speech etc. Thus, its crucial that in order to fully understand the meaning of a scene vision systems should consider data from all available modalities. This has motivated researchers to study several multi-modal problems. 
Active speaker detection is a key component in scene understanding and is an inherently multimodal (audiovisual) task. The objective here is, given a video input, to identify which person(s) are speaking in each frame. This has applications ranging from human-robot interaction to surveillance to conversational AI systems. 
%One advantage of studying this problem is its simplicity and concreteness. Its multi-modal aspect is also very clear from its definition: identification of faces(visual) generates with dialogues(audio). Furthermore, it seems like a natural stepping stone towards more challenging problems in multi-modal video analysis. \tanaya{we do not need to write these things. This is an established task/problem.}

\begin{figure}[t!]
\includegraphics[width=\linewidth]{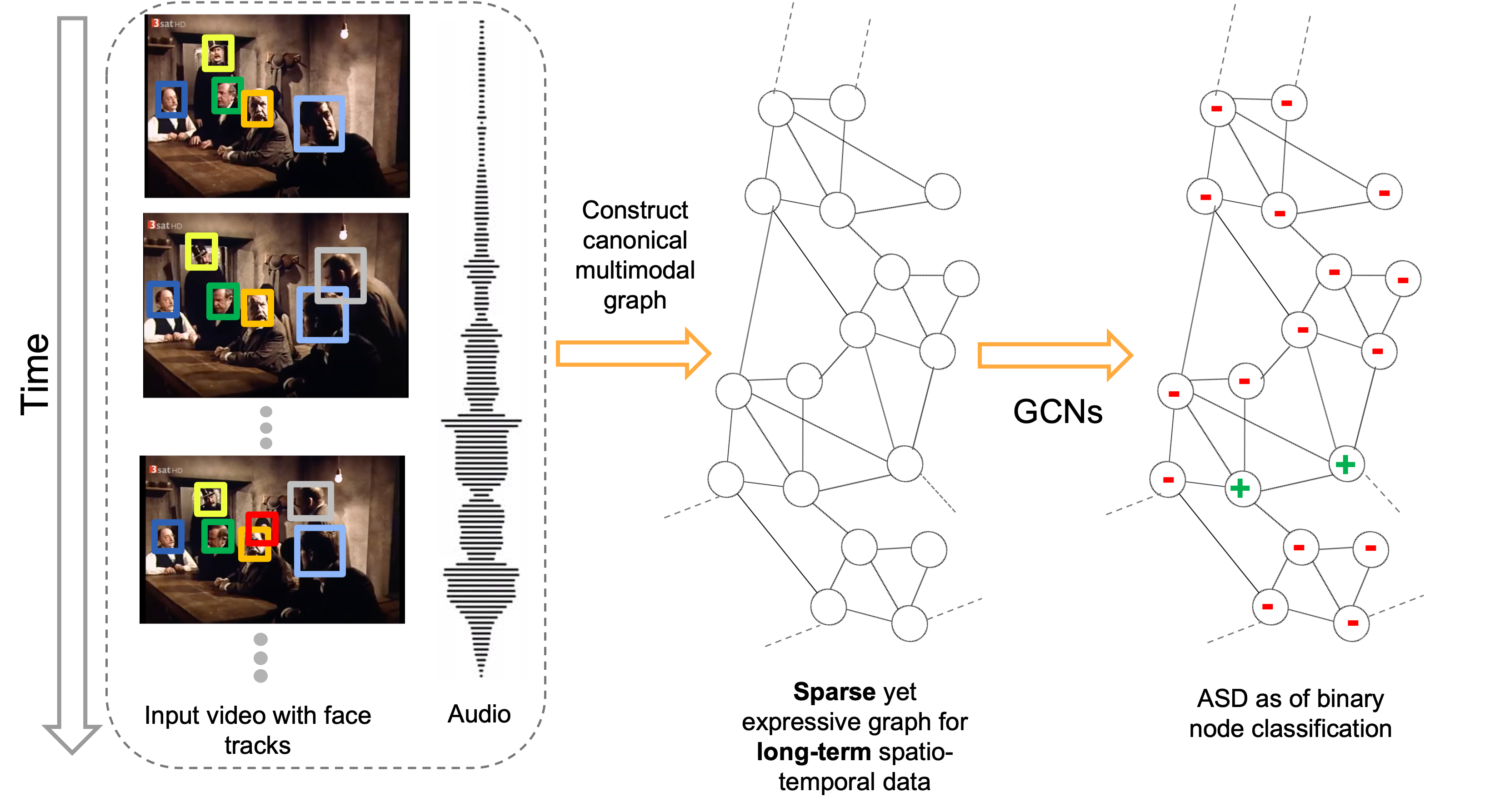}
  \caption{SPELL converts a video to a canonical graph from audio-visual input where each node corresponds to a person. The graph is dense enough for message passing across temporally-distant but relevant nodes, yet sparse enough to model their interactions within the limited memory and compute budget. The Active Speaker Detection (ASD) problem is posed as a binary node classification in this long-range spatial-temporal graph. (Best viewed in color)}
  \label{fig:teaser}
\end{figure}
%As it can be expected, 
%We note here that active speaker detection is not a new task. However, 
Earlier efforts on active speaker detection (ASD) did not meet with much success due to the unavailability of large datasets, powerful learning models or computing resources \cite{cutler2000look, everingham2006hello, everingham2009taking}. With the release of AVA-ActiveSpeaker~\cite{ava_active_speaker19} - a large and diverse ASD database, a number of promising approaches have been developed including both visual-only and audiovisual methods.
%has been made publicly available. %This dataset is large, diverse and contains long videos. Emergence of this dataset has sparked new interest in this research area. Researchers have proposed various approaches to this task including visual-only and multi-modal (audiovisual) methods. 
As visual-only methods \cite{everingham2006hello} are unable to distinguish between verbal and non-verbal lip movements, modeling audiovisual information jointly is a canonical choice for robust detection of active speakers in videos. Audiovisual approaches \cite{UniCon2021,taoSomeoneSpeakingExploring2021,alcazarActiveSpeakersContext2020,MAAS2021} address the task by extracting visual (primarily facial) and audio features from videos, following a classification of the fused multimodal features. Such models include both end-to-end trainable \cite{taoSomeoneSpeakingExploring2021} and two stage approaches \cite{alcazarActiveSpeakersContext2020,MAAS2021,UniCon2021, ASDNet_ICCV2021}. The majority of these models use large ensemble networks and complex 3D Convolutional Neural Network (CNN) features. %Recent multimodal approaches tackle the ASD task by extracting visual and audio features of the human entities in videos and then classifying the features directly after they are passed through some feature fusion(across modalities) process.%Both end-to-end trainable \cite{UniCon2021,taoSomeoneSpeakingExploring2021} and two stage model \cite{alcazarActiveSpeakersContext2020,MAAS2021} have been explored in the existing works.  

In this work, we cast active speaker detection as a graph node classification task and propose a new multimodal graph model for the same. 
First, we embed a graph on an input video where each detected face is a node and their temporal relationship is captured through edges. Next, we perform node classification (active vs inactive speaker) on this graph by learning a three layer graph neural networks (GNN) model. Graphs are chosen to naturally encode the spatial and temporal relationship (context) among the identities. It is known that GNNs can leverage such structure and provide consistent respecting neighborhood relations. Our approach, named \textbf{SPELL} (SPatial tEmporaL graph Learning), constructs a graph-based framework that is capable of modeling the temporal continuity in speech (i.e., short-term context) and also the longer term context of human conversations. Figure~\ref{fig:teaser} gives an overview of our SPELL framework. 

We note that the state-of-the-art (SOTA) models~\cite{ASDNet_ICCV2021,UniCon2021} utilize 3D CNNs, whereas SPELL uses 2D CNNs as its visual feature encoder. For example, 3DResNet-18 and 2DResNet-18 require 10 GFLOPS and 0.9 GFLOPS for fixed-point operations, respectively~\cite{ASDNet_ICCV2021}. In other words, SPELL performs comparably with SOTA models, yet requiring one order of magnitude lower computation cost. 
This computation gain comes from the unique way SPELL converts a video into a spatial-temporal graph.
The constructed graph is dense enough for message passing across temporally-distant but relevant nodes, yet sparse enough to model their interactions within a small memory and computation budget.

Our contributions are summarized below:
\begin{itemize}
    \item We propose a graph-based approach for solving the task of active speaker detection by casting it as a node classification problem. %We also strongly believe that it gives a very general framework that might be applicable to other video related tasks as well. We discuss this point in more detail in the conclusion section.
    \item Our model, SPELL, offers temporal flexibility while learning from videos by modeling short and long term temporal correlations. 
    SPELL constructs graphs that are dense enough for message passing  across  temporally-distant  but  relevant nodes,  yet sparse enough to model their interactions within tight memory constraints.
    
    % \hl{It promotes greater modularity compared to existing approaches as it can be employed on top of other available frameworks and may give non-trivial advantage.}
    %\tanaya{not clear what it means. Even if it is an advantage it should not be clubbed with temporal flexibility}% Our model provides temporal flexibility while learning from videos in the sense that it allows to model both short and long term temporal correlations. Furthermore, the proposed framework promotes greater modularization compared to many other existing approaches as it can be employed on top of other available frameworks and may give non-trivial advantage. The overall low complexity nature of 
    
    % \item Our approach offers long-range interaction modeling capability within a small memory and compute budget. SPELL constructs graphs that are dense enough for message passing  across  temporally-distant  but  relevant nodes,  yet sparse enough to model their interactions within the limited memory and compute budget. 
    
    % \item \subarna{Plug-and-play module over any feature extractor? On a second thought, we shouldn't claim it as a plug-and-play because we haven't performed experiments on other feature extractors. Perhaps, that should be our future work.}
   
   \item SPELL notably outperforms existing methods with comparable complexity on the active speaker detection benchmark dataset, Ava-ActiveSpeaker. 
   SPELL also performs comparably with SOTA models while requiring one order of magnitude lower computation cost. 
%   with significant compute and memory overhead. 

    % \item We demonstrate superior results on the AVA-ActiveSpeaker dataset outperforming several relevant and strong existing methods. 
    % Note that we did not utilize or craft any heavy machinery for feature extraction; rather used simple (off-the-self) feature extraction models, and instead focused more on building a modular model that can exploit the temporal continuity inherent in audiovisual data. 
    %Finally, we show strong results on the challenging active speaker detection data-set, AVA and superior results compared to other existing results that uses similar feature extraction procedure. Here, We want to  emphasize that main goal of this work was not to utilize or craft heavy machineries for feature extraction and performance boost. Rather, we used a very simple (off-the-self) model for feature extraction and focused more on building a modular model that can exploit the temporal continuity inherent in video data.
\end{itemize}
%\hl{name of the method: SPELL (SPatial-tEmporaL graph Learning)}
%%%%%
%% trying to come up with an abbreviation
% \subarna{Please vote for one}\\
% 1. NTeGraL
% loNg TErm GRAph Learning

% 2. INTeGraL
% Inter-modal (Intuitive) loNg TErm GRAph Learning \tanaya{+1 for INTeGraL} \kyle{+1 for INTeGraL}

% 3. ST-GRaL
% Spatial-Temporal GRAph Learning

% 4. SPELL 
% SPatio tEmporaL graph Learning \sourya{+1 for SPELL} \som{+1 for SPELL} \subarna{+1 for SPELL}

% 5. MM-SPELL
% Multi-Modal SPatio tEmporaL graph Learning

% 6. MS-GLEN
% Multimodal Spatial-temporal Graph LEarNing \kyle{+1 for MS-GLEN}

% 7. MMLT-GNN 
% Multi-modal and Longer-term GNN \kyle{+1 for MMLT-GNN}\\

% 8. SPIEGEL: 
% SPatIal-tEmporal Graph L(E)arning (has Graph in the abbreviation) 
%%%%%%%%%%%%%%%%%%%%%%%%%%%%%%%%%%%%%%%%%

\section{Related Work}
\label{sec:related}
In this section, we discuss related works in two relevant areas: application of GNNs in video scene understanding and active speaker detection.

\paragraph{GNNs for scene understanding:}
CNNs, Long Short Term Memory (LSTM) networks and their variants have long dominated the field of video understanding. In recent times, two new types of models are gaining popularity in video information processing: models based on Transformers \cite{vaswani2017attention} and GNNs \cite{}. They are not necessarily in competition with the former models, but can augment the performance of CNN/LSTM based models. Applications of specialized GNN models in video understanding include visual relationship forecasting \cite{mi2021visual}, dialog modeling \cite{geng2020spatio}, video retrieval\cite{tan2021logan}, emotion recognition \cite{shirian2020learnable} and action detection \cite{zhang2019structured}. GNN-based generalized video representation frameworks have also been proposed \cite{arnab2021unified, nagarajan2020ego,patrick2021space} that can be used for multiple downstream tasks.\\ %While, convolutional neural networks(CNN), Long Short Term Memory(LSTM) networks based models have dominated the research field of video understanding, in recent times two new existed strains of learning models are getting popular for video information processing. They are: transformer based \cite{vaswani2017attention} models and graph neural network based models. Though it must be noted that they are not necessarily in competition with the former models rather they give complementary advantage when augmented with say CNNs based models. Applications of specialized GNN based models in video understanding includes:visual relationship forcasting\cite{mi2021visual}, dialog modelling\cite{geng2020spatio}, video retrieval\cite{tan2021logan}, emotion recognition\cite{shirian2020learnable}, action detection\cite{zhang2019structured} and others\cite{nagarajan2020ego,patrick2021space}. Graph neural networks based generalized video frameworks have also been proposed\cite{arnab2021unified} that can be used across multiple tasks. 
For example, in~\cite{arnab2021unified}, a fully connected graph is constructed over the foreground nodes from video frames in a sliding window fashion, and a foreground node is connected to other context nodes from its neighboring frames.  
The message passing over the fully connected spatial-temporal graph is expensive in terms of the computational memory  and  time. Thus  in  practice  such  models  end  up using a small sliding window, making  them  unable  to  reason  over  longer-term  sequences.

SPELL also operates on foreground nodes - particularly, faces. However, the graph structure is not fully connected. We construct the graph such that it enables interactions only between relevant nodes over space and time. The graph remains sparse enough such that the longer-term context can be accommodated within a comparatively smaller memory and compute budget.    
\begin{figure*}
 \center
  \includegraphics[width=0.75\textwidth]{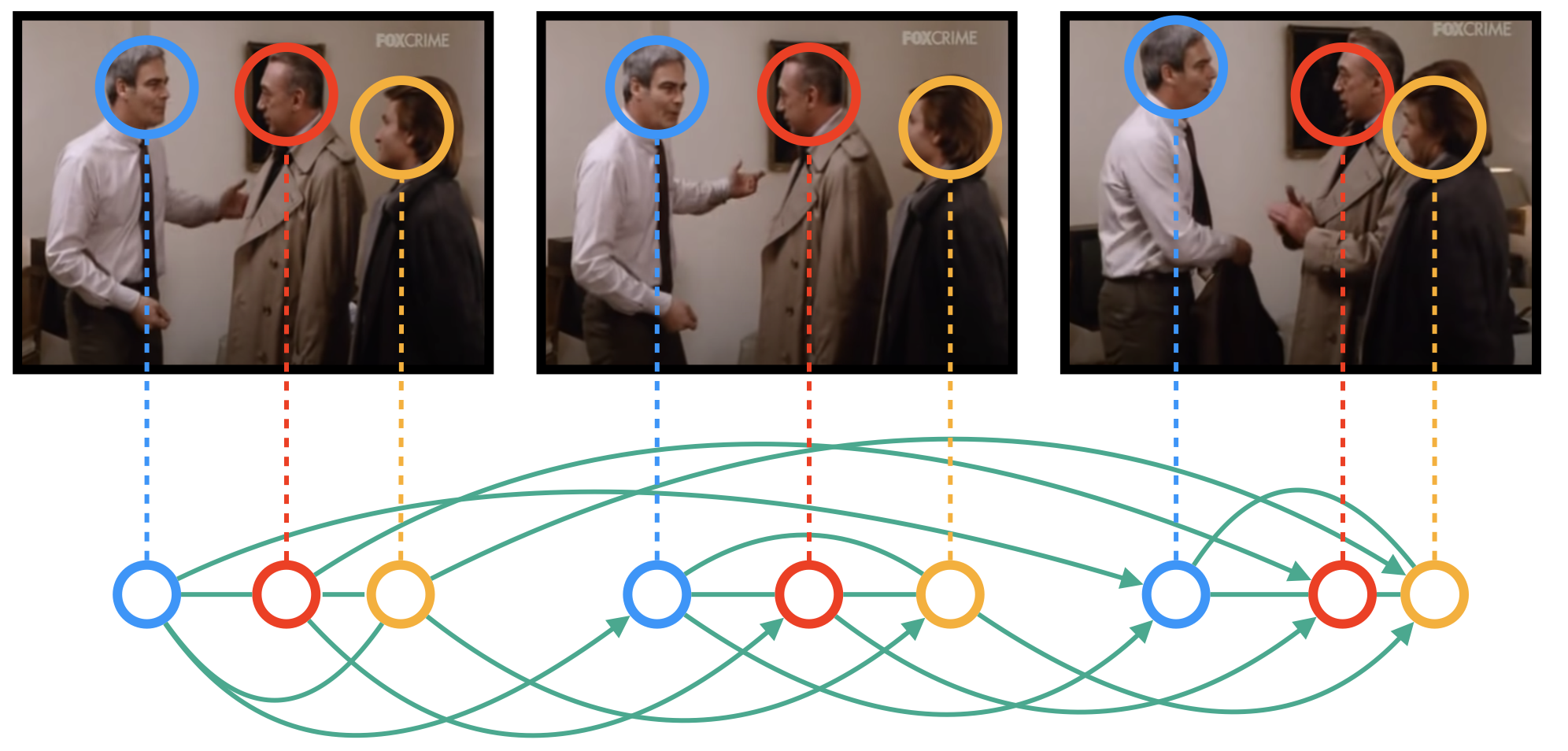}
  \caption{Forward graph embedding process.The frames above are taken from a video in the AVA dataset and they are temporally ordered from left to right.The three colors : blue, red, yellow denote three identities that are present in the frames. The graph for this portion is shown below the frames, where each node corresponds to each face in the frames. 
%   The vertices are identity color coded. 
  We connect same identities by edges across frames.
%   as they are close temporally. 
  As we can see these edges are directed and only goes in the temporally forward direction. Also, inter-identity faces are connected if they belong to the same frame. The process for creating Backward and Undirected graph(we do not show them here) embedding is identical except in the former case frame edges go in the opposite direction and the later has no directed edge. Each node also contains audio information which is not shown here. (Best viewed in color) 
%   \tanaya{I think the figure itself needs some annotation. There's space on both sides which can be used if legends are needed.}
\label{fig:graph_construction}
 }
  \end{figure*}

\paragraph{Active speaker detection (ASD):} Early work on active speaker detection by Cutler \textit{et al.} \cite{cutler2000look} detected correlated audiovisual signals using a time-delayed neural network. Subsequent works relied on visual information only considering a simpler set-up focusing on lip and facial gestures \cite{everingham2006hello}. 

The recent, high-performing ASD models rely on large networks - developed for capturing the spatial-temporal variations in audiovisual signals, often relying on ensemble networks or complex 3D CNN features \cite{alcazarActiveSpeakersContext2020, taoSomeoneSpeakingExploring2021}. Sharma \textit{et al.} \cite{Sharma2020} and Zhang \textit{et al.} \cite{Zhang2019MultiTaskLF} both used large 3D CNN architectures for audiovisual learning. The Active Speaker in Context (ASC) model \cite{alcazarActiveSpeakersContext2020} uses non-local attention modules with an LSTM to model the temporal interactions between audio and visual features encoded by two-stream ResNet-18 networks~\cite{he2016deep}. TalkNet~\cite{taoSomeoneSpeakingExploring2021} achieves superior performance through the use of a 3D CNN and a couple of Transformers~\cite{vaswani2017attention} resulting in an effectively large model. Another recent work, the ASDNet~\cite{ASDNet_ICCV2021}, uses 3D-ResNet101 for encoding visual data and SincNet~\cite{} for audio. The Unified Context Network (UniCon)~\cite{UniCon2021} proposes relational context modules to capture visual (spatial) and audiovisual context based on convolutional layers. 

Much of these advances are owing to the availability of the AVA-ActiveSpeaker database \cite{ava_active_speaker19}. Previously available multimodal databases (e.g., \cite{Chakravarty2016CrossModalSF}) were either smaller or constrained or lacked variability in data.  The work by Roth et al. \cite{ava_active_speaker19} also introduced a competitive baseline along with the large database. Their baseline model involves jointly learning an audiovisual model which is end-to-end trainable. The audio abd visual branches in this model are CNN-based which uses a depth-wise separable technique similar to MobileNets.

%\subarna{Should we add a dedicated paragraph for the dual stream network which is the backbone for Google baseline~\cite{ava_active_speaker19}, ASC~\cite{alcazarActiveSpeakersContext2020}, MAAS~\cite{MAAS2021} and our model? We have empty pages to fill up as well}
%\sourya{This section needs to be extended. Might be good idea to include classical works too as done in ASC paper}
%\hl{(recent or concurrent) can we say so? The top performers were all published in 10/21}

% Another version of UniCon, ExtendedUniCon~\cite{extendedUniCon2021}, combines various contextual information while making small changes in audio features, temporal context and loss function to achieve the best performance reported so far on the AVA-ActiveSpeaker database~\cite{ava_active_speaker19}.

The work most relevant to ours is the model called MAAS~\cite{MAAS2021} as it develops on a multimodal graph approach. Our work differs from MAAS in several ways, where the main difference in the handling of temporal context. While MAAS focuses on short-term temporal windows to construct their graphs, we focus on constructing longer-term audiovisual graphs.

\section{Methodology}
\label{sec:method}
\noindent
In this section, we describe our approach in detail. Figure ~\ref{fig:graph_construction} illustrates how SPELL constructs a graph from the input video where each node corresponds to a face within a temporal window of the video. 
SPELL is unique in terms of its canonical way of constructing the graph from a video. 
The graph is able to reason over longer-term context for all nodes without being fully-connected. The edges in the graph are only between \emph{relevant} nodes needed for message passing, leading to a sparse graph that can be accommodated within a small memory and compute budget. 
After converting the video into a graph, 
% we cast the active speaker detection problem as a binary node classification problem utilizing graph neural networks. 
we learn a light-weight GNN model to perform binary node classification on this graph. The model architecture is illustrated in Figure. \ref{fig:model}. 
The model utilizes three separate GNN modules for forward, backward and undirected graph respectively. Each module has 3 layers where the the 2nd layer is weight shared by all the three graph modules. Refer to section \ref{subsec:arch} for the intuition behind the design choice. 

%

% \tanaya{Need a short overview of our model here.} 
% We begin by reviewing graph neural networks.  \tanaya{I think we should start with 3.2, and 3.1 should be combined with either 3.2 or 3.3.}
% \subsection{Graph Neural Networks and Notations}
\subsection{Notations}
\noindent
Let $G=(V,E)$ be a graph with node set $V$ and edge set $E$. For any $v\in V$, we define $N_v$ to be the set of neighbors of $v$ in $G$. We will assume the graph has self-loops, i.e., $v\in N_v$. Let $X$ denote the set of given node features $\{\mathbf{x}_v \}_{v\in V}$ where $\mathbf{x}_v\in \mathbb{R}^d$ is the feature vector associated with the node $v$. Given this setup, we can define a $k$-layer GNN as a set of functions $\mathcal{F}=\{f_i\}_{i\in [k]}$ for $i\geq1$ where each $f_i:
V\rightarrow \mathbb{R}^m $ ( $m$ will depend on layer index $i$). All $f_i$  is parameterized by some learnable parameter. Furthermore, $X^{i}_V=\{\mathbf{x}_v\}_{v\in V}$ is the set of features at layer $i$ where  $\mathbf{x}_v=f_i(v)$ 
where we assume that $f_i$ has access to the graph $G$ and the feature set from the last layer, $X^{i-1}_V$. %We call these functions $f_i$ as simply aggregation functions (though they actually combine multiple operations in them) .
\begin{itemize}
    \item $\sf SAGE\text{-}CONV$ aggregation:
    This aggregation was proposed by \cite{graphSAGENIPS2017} and has a computationally efficient form. Given a $d$-dimensional feature set $X^{i-1}_V$%=\{\x_v\}_{v\in V}$ with all $x_v\in \mathbb{R}^d$ and
    
    , for $i\geq1$ the function $f_i:V \rightarrow \mathbb{R}^m$ is defined as follows:
    $$f(v)=\sigma \Big(\sum_{w\in  N_v}{\sf M}_i\mathbf{x}_w\Big)$$
 where, $\mathbf{x}_w\in X^{i-1}_V$, ${\sf M}_i \in \mathbb{R}^{m\times d}$ is a learnable linear transformation   and $\sigma:\mathbb{R}\rightarrow\mathbb{R}$ is a non-linear activation function applied point-wise. 
    
    \item $\sf EDGE\text{-}CONV$ aggregation:
    $\sf EDGE\text{-}CONV$~\cite{dgcnn} models pair-wise interaction between a node and its neighbors in a slightly more explicit way compared to the $\sf SAGE\text{-}CONV$ function. The aggregation function $f_i:V\rightarrow \mathbb{R}^{m}$ can be defined as:
    $$f_i(v)=\sigma \Big(\sum_{w\in  N_v}{\sf g}_i\big( \mathbf{x}_v\circ \mathbf{x}_w
 \big)	\Big)$$ where $\circ$ denotes concatenation and ${\sf g}_i: \mathbb{R}^{2d}\rightarrow \mathbb{R}^m $ is a 
learnable transformation. Often ${\sf g}_i$ is implemented by MLPs. It is easy to see that the number of parameters increases compared to $\sf SAGE\text{-}CONV$. This gives the $\sf EDGE\text{-}CONV$ layer more expressive power at a cost of reduced efficiency and possible risk of over-fitting. For our model, we set ${\sf g}_i$ to be an MLP with two layers of linear transformation and non-linearity. We describe the details in section~\ref{sec:results}.  
%\subarna{Why do we mention something that we are not using? } \\
%\subarna{What's our choice of edge function for the 1st edge-conv layer?  $H(X_i,X_j) = H(X_i, X_j - X_i)$, is it? We should probably mention that.}
    
\end{itemize}

\subsection{Video as a multi-modal graph} \label{sec:mmgraph}
\noindent
% We embed a graph on video for the active speaker detection task in a natural way. 
We represent the video as a graph suitable for the active speaker detection task.
For this we first  assume that the individual spatial locations of all faces in each frame of the input video is given. This is a valid assumption for our experimentation dataset AVA-ActiveSpeaker. Additionally, with the advent of accurate deep learning based face-detectors this assumption can be reasonably well satisfied in situations where boxes are not given. For simplicity, we assume that the entire video is represented by a single graph - if the video has $n$ faces in it the graph will have $n$ vertices. 
%This assumption is harmless and will greatly simplify the description. 
In our implementation, we temporally order the set all face-boxes in video, divide them in contiguous sets and then embed graph on each set individually.  

Let $B$ be set of all face-boxes extracted from the input video, i.e., an element $b\in B$ is a tuple $\sf(Loc,Time,Id)$ where $\sf `Loc'$ is the normalized spatial location of the face-box in its frame, `$\sf Time$' is the time-stamp of its frame and `$\sf Id' $ is a unique string which is common to all face-boxes that shares the same identity. 
Alternatively, we can identify $B$ with 
$[n]$ where $n=|B|$ is the total number of faces that appear in the video and treat 
${\sf Loc}$ as a map such that ${\sf Loc}(i):=\text{spatial location of the $i$-th face}$ for any $i\in [n]$. Similarly, ${\sf{Time}}(i)$ and ${\sf{Id}}(i)$  outputs the time and identity respectively. With this setup, the node set of  $G=(V,E)$ is set 
$V=[n]\cong B$ and for any $(i,j)\in [n]\times [n]$, we have $(i,j)\in E$ if:
\begin{itemize}
    \item  ${\sf Id}(i)={\sf Id}(j)$ and $|{\sf Time}(i)$-${\sf Time}(j)| \leq {\rm Threshold}$
    \item or, ${\sf Time}(i)={\sf Time}(j)$
\end{itemize}
where ${\rm Threshold}$ is some hyper-parameter that will be chosen in the experiments section. The intuition is that we connect two face-boxes if they share same identity and appears close in time or if they belong to same frame. To pose the active speaker detection problem as a node classification problem, we also need to specify the feature vector for each node (face-box) $v\in V$. For this, we use a pre-trained two-stream Resnet-18 as in \cite{ava_active_speaker19,alcazarActiveSpeakersContext2020} for extracting the visual features of each face-box and audio features of each frame. Then, we define feature vector of
node $v$ to be $x_v=[v_{{\sf visual}}\circ v_{{\sf audio}}]$ where $v_{{\sf visual}}$ is the visual feature of face-box $v$ from a trained two stream Resnet-18 (see below) , $v_{{\sf audio}}$ is the audio feature of $v$'s frame and $\circ$ denotes concatenation. From now onward, we write 
$G=(V,E,X)$ where $X$ is the set of node features.

\subsection{ASD as a node classification task} \label{subsec:task}
\noindent
In the previous section we described our graph embedding procedure that converts a video into a graph $G=(V,E,X)$ where each node has its own audio-visual feature vector. During the training procedure we also have access to labels of all face-boxes indicating if its an active speaker or not. Thus, the task of active speaker detection can be naturally posed as a binary node classification problem where each node(equivalently, face-box) is labeled as either speaking or not speaking. We learn a three layer graph neural network for this classification task. The first layer in the network uses ${\sf EDGE\text{-}CONV}$ aggregation and the last two layers use ${\sf SAGE\text{-}CONV}$ aggregation.
In the following section we describe a modified version of this architecture that allows us to incorporate temporal direction explicitly.

\begin{figure*}
 \center
  \includegraphics[width=\textwidth]{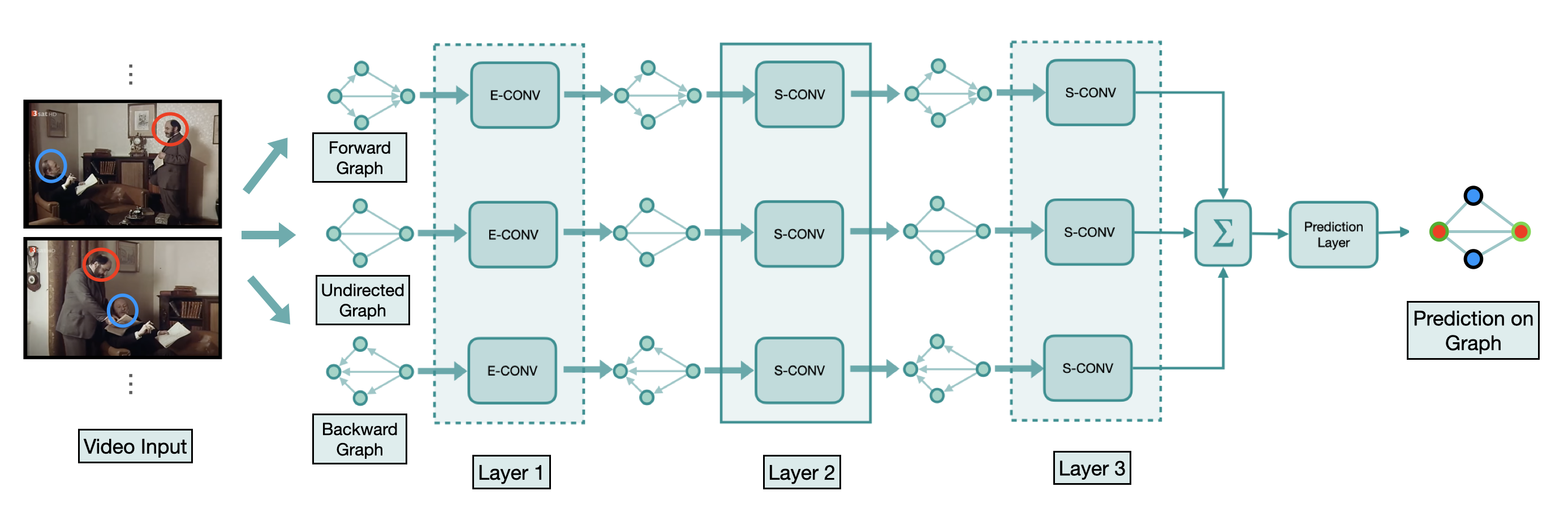}
  \caption{This figure illustrates our proposed \emph{Bi-directional} GNN model for active speaker detection. In the figure we have three separate GNN modules for Forward, Backward and undirected graph respectively. Each module has 3 layers where the the 2nd layer is weight shared by all the three graph modules. 
  We indicate this by placing the second layer inside solid lined box while for the first and third layer we use dotted lines. E-CONV and S-CONV  are shorthand for ${\sf EDGE}$-$\sf CONV$ and  $\sf SAGE$-$\sf CONV$ respectively.
  Similar to Figure~\ref{fig:graph_construction}, we use color coding: blue and red for denoting distinct identities in input frames. Furthermore,
  green and black outline denotes active speaker and non-speaker respectively in the prediction part.  The $\sum$-marked operation denotes adding 1-dimensional features (output from layer3) for each node across the three graphs.
  The prediction layer denotes  application of sigmoid to 1-dimensional feature of every node and this gives node classification probabilities.}
  \label{fig:model}
\end{figure*}

\subsection{SPELL} \label{subsec:arch}
So far, our graph embedding strategy does not consider temporal ordering into consideration. Specifically, as we use the criterion: $|{\sf Time}(i)-{\sf Time}(j)| \leq {\rm Threshold}$ for connecting vertices (face-boxes) with the same identities, the resultant graph becomes undirected. In this process we lose information about temporal ordering of the face-boxes. In order to address this issue, we augment the 
undirected graph neural network with two other parallel networks; one for going forward in time and one going backward in time. 
More precisely, in addition to the undirected graph, we create a forward graph where we connect $(i,j)$ if and only if $0\geq {\sf Time}(i)-{\sf Time}(j) \geq- {\rm Threshold}$ and similarly a backward graph that has edges only in the opposite direction. This gives us three separate graphs on which we 
learn three separate graph neural networks with weight sharing in the middle layer. Weight sharing is used in the middle layer for two reasons: parameter reduction and to enforce consistency among the responses from the different graph modules. 

In the following sections, we will refer to this augmented network as \emph{Bi-directional} or \emph{Bi-dir} for short. %\footnote{Note, this is a slight abuse of nomenclature as bi-directional often means just undirected edges. But, we will interpret it having a framework with undirected, forward-directed and backward-directed graphs as already described.
%\subarna{This notion of ``bi-directional'' is common in CV/ML community I think. 
%e.g.Bi-directional LSTM / Bi-directional GRU mean two different directional operations. 
%Do we need to add this footnote at all?}
%}
Figure~\ref{fig:model} illustrates the proposed model. We note that before we apply the graph neural network, we apply two MLP layers (linear transformation + non-linearity) to audio visual features separately and just add them to get fused features for graph vertices. After these features get passed through the first and second aggregation layers, the third aggregation layers brings down the node feature dimension to one. After this stage, for each node we simply add the three different 1D features coming from three separate graph modules and apply a sigmoid function to get the final prediction for each node.

\subsection{Feature encoding}\label{subsec:aug}
We use a two-stream 2D ResNet-18 architecture for the audio-visual feature encoder. This network is itself trained for the ASD task. We follow~ ASC \cite{alcazarActiveSpeakersContext2020} for training this model and utilize the basic outline given in AVA dataset paper \cite{ava_active_speaker19}. The network takes visual input in terms of a stack of $k$ face-crops(or face-boxes) from some time interval ${\sf t}$ and audio input as Mel-spectrogram of the audio-wave extracted from this time interval. These two different modalities go into two different sub-networks. In the end, the audio and visual features, that are output of these two sub-networks, are combined via concatenation and then then the concatenated embedding is passed through a prediction layer to get the final prediction.

\paragraph{Data augmentation} Reliable ASD models should be able to detect speaking signals even if there is a noise in the audio. To make our method robust to noise, we make use of data augmentation methods while training the feature extractor. %First, we apply SpecAugment~\cite{Park2019SpecAugmentAS}, which was originally proposed for the task of speech recognition. Specifically, we use two randomly-generated masks to zero out some of the consecutive time and frequency bands of the audio spectrogram.
Inspired by TalkNet~\cite{taoSomeoneSpeakingExploring2021}, we augment the audio data by negative sampling. For each audio signal in a batch, we randomly select another audio sample from the whole training dataset and add it after decreasing its volume by a random factor. This technique can effectively increase the amount of training samples for the feature extractor by selecting negative samples from the whole training dataset.

\paragraph{Spatial feature} We also incorporate the spatial features corresponding to each face as additional input to the node feature. We project the 4-D spatial feature of each face region parameterized by the normalized center location, height and width to a 64-D feature vector using a single fully-connected layer. The resulting spatial feature vector is then concatenated to the visual feature at each node.

%parameterized by the normalized location and size corresponding to each face box are also utilized in the gnn layers.

% \paragraph{Self-supervised Learning}
% Audio Video sync as pretext task
% \subarna{Delete this paragraph if pretraining with SS doesn't work}

\section{Experiments}
\label{sec:results}
\paragraph{Dataset}

We perform experiments on the large-scale AVA-ActiveSpeaker\footnote{\url{https://research.google.com/ava/download.html\#\#ava\_active\_speaker\_download}} dataset~\cite{ava_active_speaker19}. This dataset is derived from Hollywood movies, and face tracks are provided. The AVA dataset provides extensive annotations of the available face tracks, a key feature that was missing in its predecessors. 
% Moreover, being a dataset that is built upon various movies, it offers rich diversity of data samples. 
The training, validation and test split consists of $29,723$, $8,015$ and $21,361$ video utterances respectively, each ranging from 1 to 10 seconds.   

\paragraph{Implementation details}
Following~\cite{ava_active_speaker19}, we utilize a two-stream network with a ResNet-18~\cite{he2016deep} backbone pre-trained on ImageNet~\cite{imagenetdeng2009}.
We perform visual augmentation as in~\cite{ava_active_speaker19} and audio augmentation as described in Section~\ref{subsec:aug}. 
% following~\cite{taoSomeoneSpeakingExploring2021}. 
% \subarna{if the newest spec-augment works - we will update the above text}.
We train the network end-to-end first with cross-entropy loss. We follow the supervision strategy outlined in~\cite{ava_active_speaker19} and train the model to learn useful features from both the video and audio streams. Both audio and video features from the two stream network are 512 dimensional. 

% Next, we extract the vertex features $x_v$ using the two-stream feature encoder of ASC~\cite{alcazarActiveSpeakersContext2020}, which is based on ResNet-18 architecture~\cite{he2016deep}. This encoder is trained on the sequences of face crops and the Mel-spectogram of audio signal. Then, we build our graph neural network that works on the pairs of audio and visual features, which are extracted by the encoder. Specifically, we implement our network by using PyTorch Geometric library~\cite{pytorchgeometric2019}. We want to note that our network is a lightweight one (0.8MB).

\paragraph{Training SPELL} 
We implement SPELL using PyTorch Geometric library~\cite{pytorchgeometric2019}. Our network model contains 3 GCN layers, each with 64 dimensional filters. The first layer is an $\sf EDGE$-$\sf CONV$ layer where 
the aggregation function utilizes a two layer MLP for feature projection.
The second and third GCN layers are of type $\sf SAGE$-$\sf CONV$ and each of them uses a single MLP layer.  

% Next, we extract the vertex features $x_v$ using the two-stream feature encoder of ASC~\cite{alcazarActiveSpeakersContext2020}, which is based on ResNet-18 architecture~\cite{he2016deep}. This encoder is trained on the sequences of face crops and the Mel-spectogram of audio signal. 
We use the encoded audio, visual, and spatial features for each face-box, then we build our graph neural networks that uses those as node features. We set the number of nodes 
to $n$ to 2000 and set $\mathrm{Threshold}$ parameter to 0.9 seconds. In the ablation section, we call $\mathrm{Threshold}$ as `Time $\mathrm{Threshold}$' to be more explicit. 
% , which are extracted from the encoder. 
% We want to note that our network is a lightweight one (0.8MB).
We train our graph neural network with a batch size of 16 using the Adam optimizer~\cite{kingma2014adam}. The learning rate starts at $2 \times 10^{-4}$ and decays following the cosine annealing schedule~\cite{loshchilov2016sgdr} with 10 maximum iterations. The whole training process of 120 epochs takes less than two hours using a single GPU (TITAN V).

%\subarna{Use positional feature~\cite{LAEO-Net_Marin_JimenezKM19} ? - UniCon~\cite{UniCon2021} used them. Concatenating spatial feature (late fusion) with audio-visual feature gave them 1.7 point boost. } \\
% Table 3 in UniCon \url{https://arxiv.org/pdf/2108.02607.pdf} Non-Active Speaker suppression ?? \\

\begin{table}[b!]
\centering
\caption{Comparison with the state-of-the-art on the AVA-ActiveSpeaker 
validation set 
in terms of mean average precision (mAP). $\dagger$ 
% $\star$ 
denotes 3DResNext-101 and 3DResNet-18 as their corresponding visual encoders. Top section of the table corresponds to the models that use 2DResNet-18 as the visual encoder backbone. $\star$ Stage-1 mAP of UniCon~\cite{UniCon2021} is higher than ~\cite{alcazarActiveSpeakersContext2020, ava_active_speaker19}.
}
\vspace{2mm}
\setlength{\tabcolsep}{15pt}
\resizebox{1\linewidth}{!}{%
\begin{tabular}{l c}
\toprule
\textbf{Method} & \textbf{mAP (\%)} \\
% \midrule
% \multicolumn{2}{c}{\emph{Validation set}}\\
\midrule
AVA-ActiveSpeaker baseline~\cite{ava_active_speaker19} & 79.2 \\
% Sharma \textit{et al.}~\cite{Sharma2020} & 82.0 \\
Zhang \textit{et al.}~\cite{Zhang2019MultiTaskLF} & 84.0 \\
ASC-temporal context \cite{alcazarActiveSpeakersContext2020} & 85.1\\
MAAS-LAN~\cite{MAAS2021} & 85.1 \\
Chung et al. \cite{Chung2019} & 85.5 \\
ASC~\cite{alcazarActiveSpeakersContext2020} & 87.1 \\ 
MAAS-TAN~\cite{MAAS2021} & 88.8 \\
\textbf{SPELL} (Ours) & 90.6 \\
UniCon~\cite{UniCon2021}$\star$ & 92.0 \\
\midrule
% TalkNet~\cite{taoSomeoneSpeakingExploring2021} & 92.3 \\
% Sharma \textit{et al.}~\cite{Sharma2020} & 82.0 \\
% UniCon~\cite{UniCon2021}$^\star$ & 92.0 \\
ASDNet~\cite{ASDNet_ICCV2021}$^\dagger$ & \textbf{93.5} \\

% \midrule
% \multicolumn{2}{c}{\emph{Test set}}\\
% \midrule
% AVA-ActiveSpeaker baseline~\cite{ava_active_speaker19} & 82.1 \\
% ASC~\cite{alcazarActiveSpeakersContext2020} & 86.7 \\ 
% Chung et al. \cite{Chung2019} & 87.8 \\
% MAAS-TAN~\cite{MAAS2021} & 88.3 \\
% \textbf{SPELL} (Ours) & \\

\bottomrule
\end{tabular}
\label{tab:main_table}
}
\end{table}

\subsection{Comparison with SOTA}

% We evaluate the performance using Sklearn library \footnote{https://scikit-learn.org/stable/modules/generated/sklearn.metrics.f1\_score.html}
Results on the AVA-ActiveSpeaker validation set is summarized in Table~\ref{tab:main_table}. 
The models in the bottom section of the table utilize compute-heavy 3D CNN visual feature encoders, while those in the top section use significantly lighter 2D CNN feature encoders. 
Performances of all methods are evaluated by mean average precision (mAP) metric using the official tool from ActivityNet. 
% \footnote{\url{https://github.com/activitynet/ActivityNet}}. 

A concurrent and closely related work MAAS~\cite{MAAS2021} also uses graphs. While  MAAS-LAN includes a single timestamp and 4 videos, MAAS-TAN consists of 13 temporally linked local graphs spanning about 1.59 seconds. Unsurprisingly, MAAS-TAN that spans a longer temporal window compared with MAAS-LAN, performs better in terms of mAP metric. Also note that, SPELL performs single forward pass, whereas MAAS performs multiple forward passes during inference. 
% We note that SPELL spanning a temporal window of f0.9 seconds outperforms MAAS-TAN. 
% % spanning lower temporal window (0.9 seconds vs 1.59 seconds) % already outperforms MAAS-TAN by large margin due to SPELL's capability of capturing the longer term context. 
The Table~\ref{tab:main_table} shows that SPELL outperforms all models including MAAS with comparable complexity that use 2D CNN backbone, and performs comparably with compute-heavy SOTA methods that use an order of magnitude higher complexity 3D CNN feature encoders.

\begin{table}[b!]
\centering
\caption{Contextual reasoning capacity. We quantify the contextual reasoning capacity by the improvement of mAP over the corresponding feature-encoding stage. 
ASDNet$\dagger$ uses 3DResNext-101 and SincDSNet for video and audio feature, respectively; 
% UniCon$\star$ uses 3DResNet-18 and 2DResNet-18 for video and audio encoding, respectively. 
Others use 2DResNet-18 for both audio and video feature encoding. All methods are two-stage methods except ASDNet which is a three-stage model.}
\vspace{2mm}
\setlength{\tabcolsep}{5pt}
\resizebox{1\linewidth}{!}{
\begin{tabular}{l ccc}
\toprule
\textbf{Method} & \textbf{Stage-1 mAP} & \textbf{Final mAP} & \textbf{$\Delta$mAP} \\
\midrule
MAAS-LAN~\cite{MAAS2021} & 79.2 & 85.1 & 5.9\\
ASC~\cite{alcazarActiveSpeakersContext2020} & 79.2 & 87.1 & 7.9\\ 
Unicon~\cite{UniCon2021} & 84.0 & 92.0 & 8.0 \\
ASDNet~\cite{ASDNet_ICCV2021}$\dagger$ & \bf 88.9$\dagger$ & \bf 93.5 & 4.6\\
MAAS-TAN~\cite{MAAS2021} & 79.6 & 88.8 & 9.2\\
% TalkNet~\cite{taoSomeoneSpeakingExploring2021} & 92.3 \\
\textbf{SPELL (Ours)} & 80.2 & 90.6 & \textbf{10.4} \\
\bottomrule
\end{tabular}
\label{tab:context_capacity}
}
\end{table}

\subsection{Contextual reasoning capacity}
Most ASD models use multi-stage training and utilize dedicated modules to further improve the performance over feature encoders by exploiting long-term context. We refer to all such dedicated modules by the generic term 
`contextual reasoning module'. SPELL (excluding the feature encoding module) is one such contextual reasoning module. 
Empirically, we can quantify the capacity of the contextual reasoning modules by how much gain they can provide atop their corresponding baseline feature encoders. Table \ref{tab:context_capacity} compares SPELL with other contextual reasoning modules proposed in the state-of-the-art models. 
Methods such as 
% UniCon~\cite{UniCon2021} and 
ASDNet~\cite{ASDNet_ICCV2021} that use 3D CNN for video feature encoding, implicitly incorporate temporal context within the feature encoder unlike methods using 2D CNN based feature encoders. 
We note that the backbone audio and video feature encoders for ASC, MAAS and SPELL are the same - 2D ResNet-18. For SPELL and MAAS, the contextual reasoning modules are both based on graphs. 
MAAS-TAN uses longer temporal context unlike MAAS-LAN, and thus the former shows larger performance gain over the feature encoding baseline. 
Table~\ref{tab:context_capacity} shows that SPELL outperforms all methods in terms of contextual reasoning capacity.

\begin{table}[tb]
\Large
\caption{Comparison of model complexity: Model size excludes the feature encoder which is a 2D ResNet-18 for each of the video and audio modalities for all methods in the table.$\dagger$ denotes estimated number of parameters, and model size.}
\vspace{2mm}
\label{tab:modelsize}
% \vspace{-1mm}
% \renewcommand*{\arraystretch}{1.4}
\setlength{\tabcolsep}{10pt}
\resizebox{1\linewidth}{!}{%
\begin{tabular}{cccc}
\toprule
{\bf Model}  & {\bf $\#$ of Params} & {\bf Size} (MB) & \bf{mAP (\%)}\\ 
% \hline \hline
\midrule
ASC \cite{alcazarActiveSpeakersContext2020} & 1,122,560 & 4.5  & 87.1  \\
MAAS-TAN \cite{MAAS2021}$^\dagger$ & 131,200 & 0.56   & 88.8 \\
\textbf{SPELL (Ours)} & 111,808 & 0.48  & 90.6 \\
\bottomrule
\end{tabular}
}
\end{table}

\subsection{Model efficiency}
In Table~\ref{tab:modelsize}, we compare the complexity of our model with that of others sharing the same feature encoder architecture.
% For ASC, we directly use their code repository to compute the number of parameters and size. 
We estimate the number of parameters and the model size for MAAS-TAN 
from the model description in the paper since neither the source code nor the model is available. ASC has a $\sim10$x larger model compared to ours - however, our model achieves  $3.1\%$ higher mAP compared to ASC. Our model has fewer number of parameters than MAAS-TAN as well while achieving $1.4\%$ higher mAP. To summarize, SPELL outperforms all existing methods with comparable complexity. 
Notably, 3D ResNet-18 and 2D ResNet-18 require $10 GFLOPS$ and $0.9 GFLOPS$ for fixed-point operations respectively~\cite{ASDNet_ICCV2021}. 
Thus, SPELL is one order of magnitude more computationally efficient than state-of-the-art methods that utilize 3D CNNs but achieves similar performance as those methods.

In Table~\ref{tab:inftime}, we show the total inference time (excluding the feature extraction time) on the entire validation set for ASC~\cite{alcazarActiveSpeakersContext2020} and SPELL run on the same environment. We can see that SPELL is almost four times faster than ASC. For fair comparison(batching scheme is not comparable between ASC and SPELL), we do not report per sample inference time.

\begin{table}[tb]
\small
\caption{Comparison of the total inference time (excluding the feature extraction time) measured on Titan V on the Validation set.}
\vspace{2mm}
\label{tab:inftime}
% \vspace{-1mm}
% \renewcommand*{\arraystretch}{1.4}
\setlength{\tabcolsep}{25pt}
\resizebox{1\linewidth}{!}{%
\small
\begin{tabular}{cc}
\toprule
{\bf Model}  & {\bf Inference time}\\ 
% \hline \hline
\midrule
ASC \cite{alcazarActiveSpeakersContext2020} & 11,972 ms  \\
\textbf{SPELL (Ours)} & 3,272 ms \\
\bottomrule
\end{tabular}
}
\end{table}

\begin{figure}
 \center
  \includegraphics[width=0.45\textwidth]{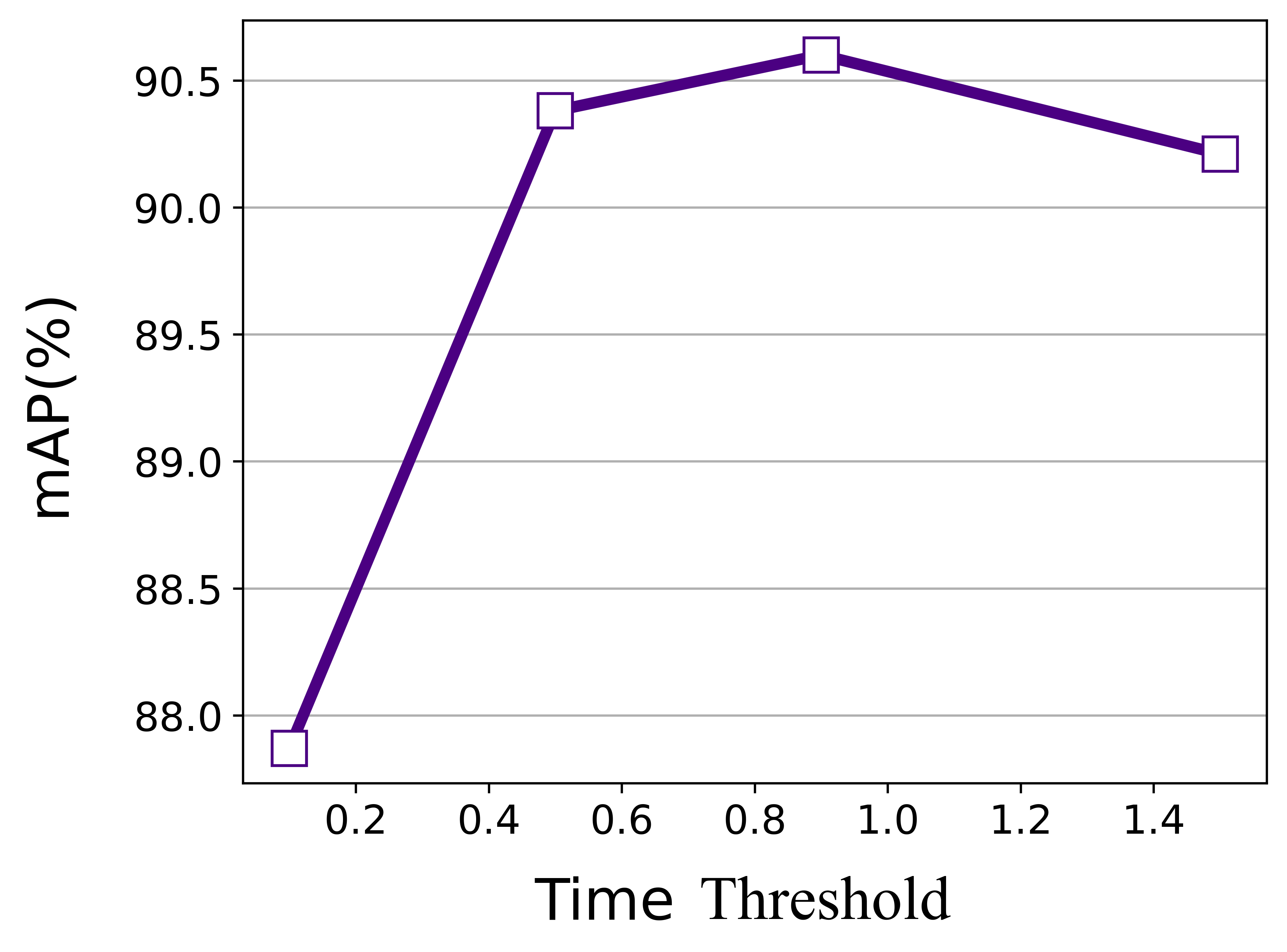}
  \includegraphics[width=0.45\textwidth]{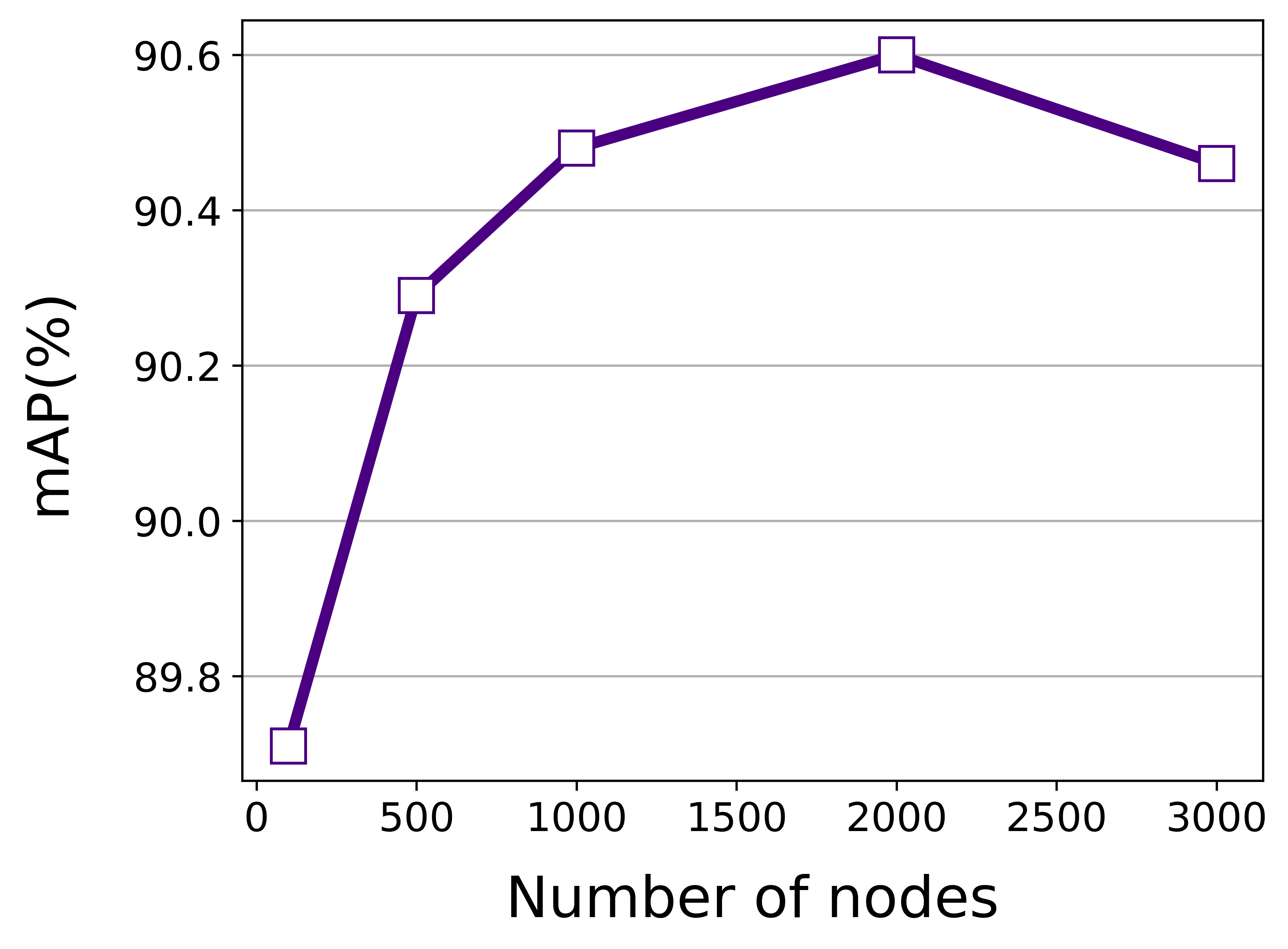}
  \caption{Study on the impact of two hyperparameters, which are Time {$\rm  {Threshold}$} (when number of vertices, $n$, is set to 2000) and number of vertices (with 0.9  Time $\rm{Threshold}$).
  %\subarna{Use 'Number of nodes' in the X-axis instead of 'n'} \kyle{I will upload the new one after our last meeting see you soon}
  }
  \label{fig:4}
\end{figure}
\subsection{Ablation experiments}
We ablate SPELL to validate the individual contribution of each design choice, namely graph structure (unidirectional vs bi-directional), audio data augmentation, spatial-features, and edge drop out. We summarize our ablation results in Table~\ref{tab:ablation}. We can observe that the bi-directional graph structure and audio data augmentation play significant roles in boosting the detection performance. This implies that the information about temporal ordering of the face boxes is important and that the negatively sampled audio makes our model more robust to the noise. Additionally, edge drop out and spatial encoding also bring meaningful performance gains.

\begin{table}[tb]
\caption{Ablation study on algorithmic modules evaluated on the AVA-ActiveSpeaker val set.
}
\vspace{2mm}
\label{tab:ablation}
% \vspace{-1mm}
% \renewcommand*{\arraystretch}{1.4}
\setlength{\tabcolsep}{8pt}
\resizebox{1\linewidth}{!}{%
\begin{tabular}{ccccc c}
\toprule
{\bf Graph}  & {\bf Bi-dir} & {\bf Aug} & {\bf Drpt} & {\bf SP-feat} & {\bf mAP} (\%) \\ 
% \hline \hline
\midrule
- & - &-   &- & -& 79.2     \\ % 79.2
- & - &\checkmark & - & -& 80.3    \\ % 80.25
\checkmark & - &-   &- & -& 87.6     \\ % 87.62
\checkmark & \checkmark &-   &- & -& 88.1     \\ % 88.08
\checkmark & -  &\checkmark & - & -& 89.8    \\ % 89.81
% \checkmark & - &-   &\checkmark & -     \\
\checkmark & \checkmark &\checkmark   &  -& &90.1   \\ % 90.13
\checkmark & \checkmark &   \checkmark & \checkmark &- &90.2\\ % 90.22
\checkmark & \checkmark &   \checkmark & \checkmark & \checkmark & 90.6\\ % 90.56
\bottomrule
\end{tabular}
}
\end{table}

% Number of nodes and temporal context window are treated as hyperparamaters in our model. 
% \subarna{describe these two tables or plots}

\begin{table}[tb]
\caption{Ablation on input modalities evaluated on the AVA-ActiveSpeaker val set.}
\vspace{2mm}
\label{tab:modality}
% \vspace{-1mm}
% \renewcommand*{\arraystretch}{1.4}
\setlength{\tabcolsep}{25pt}
\resizebox{1\linewidth}{!}{%
\begin{tabular}{cc c}
\toprule
{\bf Video}  & {\bf Audio} & {\bf mAP} (\%) \\ 
% \hline \hline
\midrule
- &\checkmark & 59.5    \\
\checkmark & -& 75.3     \\
\checkmark & \checkmark & 90.6     \\
\bottomrule
\end{tabular}
}
\end{table}

\begin{figure*}
 \center
  \includegraphics[ width=\textwidth, ]{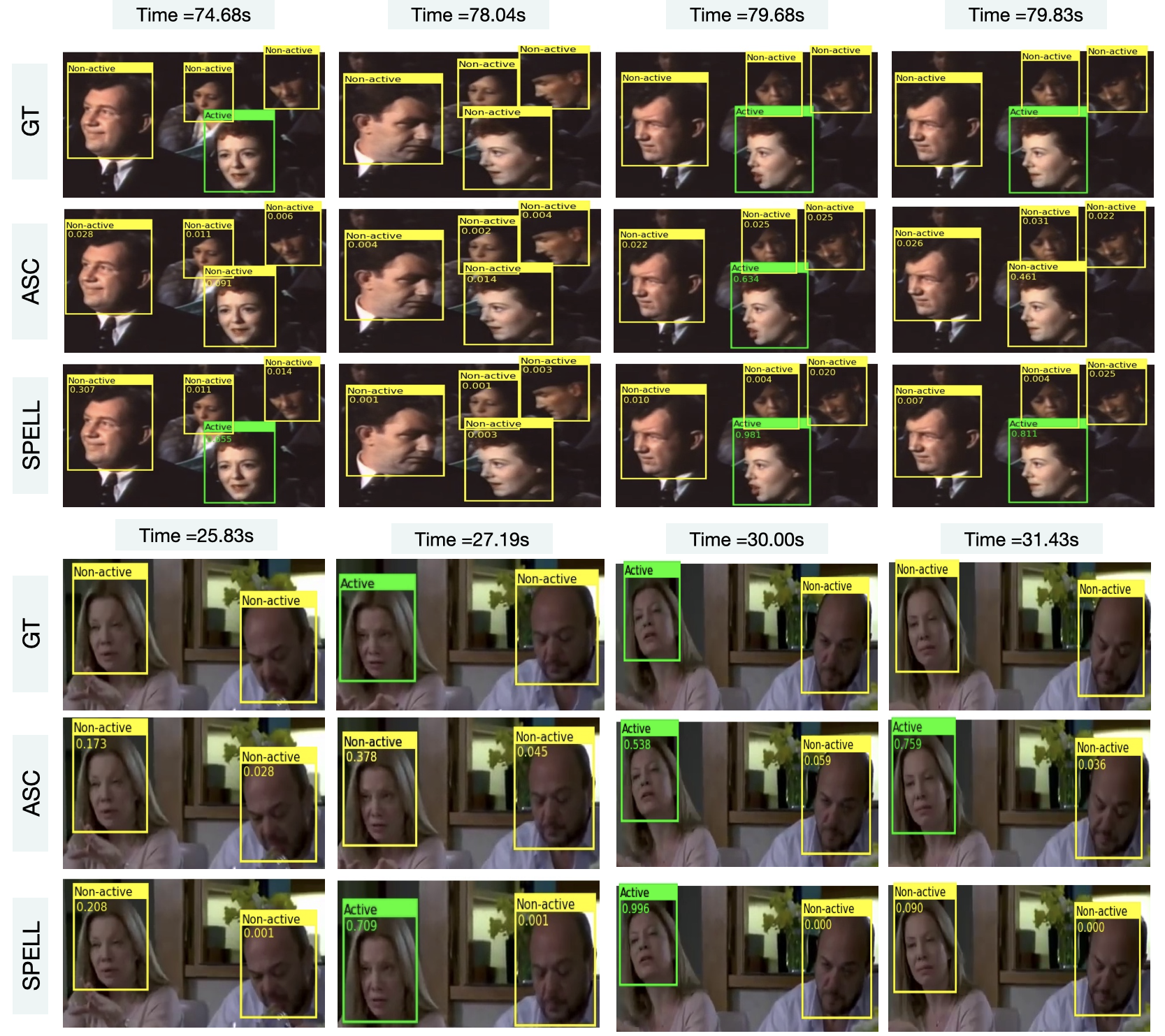}
  \caption{Qualitative results:  The green bounding boxes indicate the active speakers and yellow indicates non-active speaker. Frame sequences are taken over 5-6 seconds interval. GT: Ground Truth.
%   \tanaya{Can we write 'ASC [2]' and 'SPELL (Ours)'?}
  }
  \label{fig:quality}
\end{figure*}
In Figure~\ref{fig:4}, we analyze the impact of two hyperparameters: Time ${\rm  Threshold}$ (Section~\ref{sec:mmgraph}) and number of vertices in the embedded graph. Time ${\rm Threshold}$ controls the  (Section~\ref{sec:mmgraph}) connectivity or the edge density in the embedded graph. Specifically, larger values for Time ${\rm Threshold}$ allow us to model longer temporal correlations but increases the average degree of the vertices, thus making the system more computationally expensive. Moreover, from Figure~\ref{fig:4},
we can observe that connecting faces that span over 1 sec deteriorates the detection performance.
One potential reason behind this could be that larger degree results in over averaging during the aggregation procedure in the graph. Interestingly, we also found that a larger number of vertices does not always lead to higher performance. 
This might be because after a certain threshold number of vertices the degree distribution tends not to change much. %\subarna{complete}
%\kyle{@sourya help me with this paragraph; and please let me know if the labels of Fig~\ref{fig:4} are incorrect so that I can redraw them}

In Table~\ref{tab:modality}
we report on an ablation study involving audio and video features. As expected, our approach achieves best performance when both the modalities are used. Detection performance drops significantly, from $90.22\%$ mAP score to $59.66\%$ mAP, in the audio feature only case; while the performance drop in the video only case is relatively lower. These experiments clearly demonstrate the importance of including both audio and video modalities for this application. 

\subsection{Qualitative results}
In Figure~\ref{fig:quality}, we show several detection examples to demonstrate qualitative performance of our approach. We show results on frames with many faces, spanning moderately long interval of 5-6 seconds, sometimes without any active speaker at all.  
In the first sequence, while SPELL correctly finds the active speaker in the first time-stamp, ASC incorrectly labels all the faces as non-active speaker with high confidence. ASC makes similar error in last time-stamp of sequence one and second time stamp of second sequence. We show an example of   
false positive detection by ASC. In last time-stamp of second sequence, we can see that there is no speaker present. For this, SPELL outputs correctly and confidently. However, ASC
falsely detects an active speaker with high confidence. 

% We have successfully applied this framework for binary classification problem of Active Speaker detection. Although we believe SPELL can act as a generic node classification problem; we are yet to validate its efficacy for other generic multi-class node classification problem. 

\section{Discussion}
\label{sec:discussions}

% Our method aims to optimize a graph that simultaneously detects 
\paragraph{Limitations and future work:}
In this section, we discuss some of the potential limitations of our proposed approach. 
Our canonical and unique graph structure relies on input face tracks that define the temporal connections among nodes in the graph. SPELL has yet to be tested on in-the-wild dataset where ground truth face tracks are not provided.  
Another issue with SPELL is that it is not an end-to-end trainable system.
% which is 
% a common feature shared by 
% % inherent in
% almost all graph neural network based models. 
% Two points must be noted here. 
% First, this 
This \emph{issue} is in some sense a double edged sword. On one hand as it is a feature extractor agnostic approach that can be easily augmented with any other module. On the other hand, we may get poor performance due to the use of bad feature extractors.
% Secondly, it is certainly possible to transform our(or any other GNN based approach) algorithm into an end-to-end system. However that will incur significant amount of memory and computation. 

\paragraph{Conclusion:}
We proposed a canonical graph based approach towards active speaker detection in videos. The basic idea is to capture the spatial and temporal relationship among and across  identities through a graph structure while being aware of the long-term temporal dependencies in data. Our compact model offers simplicity and modularity without compromising performance accuracy. We achieve high accuracy, outperforming strong baselines, improving over the only other graph-based approach (i.e., MAAS-TAN) and producing comparable results to the current state-of-the-art (3D CNNs) with a significantly smaller model. The model we propose is also generic; it can be used to address other multimodal tasks such as action recognition.

% \section*{Appendix}
% \label{sec:supp}
% \input{supplementary}

{\small
\bibliographystyle{ieee_fullname}
\bibliography{main}
}

\end{document}

% \author[1]{XXX}
% \author[2]{YYY}
% \author[1]{ZZZ}
% \affil[1]{University of .... }
% \affil[2]{Intel Labs, USA}
% \renewcommand\Authands{, and }
\author{First Author\\
Institution1\\
Institution1 address\\
{\tt\small firstauthor@i1.org}
\and
Second Author\\
Institution2\\
First line of institution2 address\\
{\tt\small secondauthor@i2.org}
}
\maketitle

\appendix
\section{Network architecture details}
For clarification, network architecture of SPELL is shown in Table~\ref{tab:arch}.

\begin{table}[htbp]
\centering
\caption{Detailed architecture of SPELL. Batch normalization~\cite{ioffe2015batch} and ReLU~\cite{nair2010rectified} follow after each of (6), (7), and (9-14).}
\vspace{2mm}
\resizebox{1\linewidth}{!}{%
\begin{tabular}{c|c|c|c}
\toprule
\textbf{Index} & \textbf{Inputs} & \textbf{Description} & \textbf{Dimension} \\
\midrule
(1) & - & 4-D spatial feature & 4 \\
(2) & - & Visual feature $v_{{\sf visual}}$ & 512 \\
(3) & - & Audio feature $v_{{\sf audio}}$ & 512 \\
(4) & (1) & Linear (4 $\rightarrow$ 64) & 64 \\
(5) & (2), (4) & Concatenation & 576 \\
(6) & (5) & Linear (576 $\rightarrow$ 64) & 64 \\
(7) & (3) & Linear (512 $\rightarrow$ 64) & 64 \\
(8) & (6), (7) & Addition & 64 \\
(9) & (8) & $\sf EDGE\text{-}CONV$ (Forward) & 64 \\
(10) & (8) & $\sf EDGE\text{-}CONV$ (Undirected) & 64 \\
(11) & (8) & $\sf EDGE\text{-}CONV$ (Backward) & 64 \\
(12) & (9) & $\sf SAGE\text{-}CONV$ (Forward, Shared) & 64 \\
(13) & (10) & $\sf SAGE\text{-}CONV$ (Undirected, Shared) & 64 \\
(14) & (11) & $\sf SAGE\text{-}CONV$ (Backward, Shared) & 64 \\
(15) & (12) & $\sf SAGE\text{-}CONV$ (Forward) & 1 \\
(16) & (13) & $\sf SAGE\text{-}CONV$ (Undirected) & 1 \\
(17) & (14) & $\sf SAGE\text{-}CONV$ (Backward) & 1 \\
(18) & (15), (16), (17) & Addition & 1 \\
(19) & (18) & Sigmoid & 1 \\

\bottomrule
\end{tabular}
\label{tab:arch}
}
\end{table}

\section{Additional experiments}

We perform additional experiments with different filter dimensions of the GCN layers. In Table~\ref{tab:fdim}, we show how the detection performance and the model size change depending on the filter dimension. We can observe that increasing the filter dimension above 64 does not bring any performance gain when the model size increases significantly. Therefore, as described in our main paper, it was chosen to use GCN layers with 64 dimensional filters in SPELL.

In addition, we perform an experiment with a different network structure. As illustrated in Figure 3 of our main paper, the weight of SPELL's 2nd layer (Layer 2) is shared across all the graph modules. We show that increasing the capacity of this layer can bring the performance gain. Inspired by the recent success of Graph Inception Network~\cite{shirian2021dynamic}, we adopt the idea of graph inception layer in the 2nd layer of SPELL. Specifically, it consists of three parallel $\sf SAGE\text{-}CONV$ operations each with 16, 32, and 64 feature dimensions and a single 1-hop maxpool operation. The aggregated features by $\sf SAGE\text{-}CONV$ and the pooled features are concatenated and passed to the 3rd layer of SPELL. This modification boosts the score to 90.7 mAP (\%), but the model size also increases to  0.78 MB.

\begin{table}[htbp]
\centering
\caption{Comparisons of the detection performance and the model size with different filter dimensions on the AVA-ActiveSpeaker dataset~\cite{ava_active_speaker19}.}
\vspace{2mm}
\setlength{\tabcolsep}{18pt}
\resizebox{1\linewidth}{!}{%
\begin{tabular}{c|c|c}
\toprule
\textbf{Filter Dim} & \textbf{mAP (\%)} & \textbf{Size (MB)} \\
\midrule
16 & 89.7 & 0.10 \\
32 & 90.4 & 0.20 \\
\textbf{64} & \textbf{90.6} & \textbf{0.48} \\
128 & 90.4 & 1.3 \\
256 & 90.2 & 3.9 \\
\bottomrule
\end{tabular}
\label{tab:fdim}
}
\end{table}

\section{Additional qualitative results}
We provide more qualitative results of SPELL. Please refer to the videos that are included in the subfolder.

\section{Source code}
Our source code is available here: 
%\hl{if the anonymized link works, or just mention it's added as a part of the supplementary material}
 \url{https://anonymous.4open.science/r/SPELL-E6F0}

%\begin{figure*}[htbp]
%\centering
%  \small
%    \includegraphics[width=0.85\linewidth]{images/A1.png}\\[-0.8ex] \hspace*{14em}(a) \\[1ex]
%    \includegraphics[width=0.85\linewidth]{images/A2.png}\\[-0.8ex] \hspace*{16em}(b) \\[1ex]
%    \includegraphics[width=0.85\linewidth]{images/A3.png}\\[-0.8ex] \hspace*{18em}(c)
%  \caption{asaa}
%  \label{fig:addqual}
%\end{figure*}

{\small
\bibliographystyle{ieee_fullname}
\bibliography{main}
}